\documentclass{article}
\usepackage{spconf,amsmath,graphicx}
\usepackage{amssymb,amsmath,epsfig,cite,url,multicol,multirow}
\usepackage{amsmath,graphicx}
\usepackage{xcolor}
\usepackage{hyperref}
\usepackage{graphicx}
\usepackage{float}
\usepackage{comment}
\usepackage{float}
\usepackage{diagbox,balance}
\floatstyle{plaintop}
\restylefloat{table}

\title{Exploring Teacher-Student Learning Approach for Multi-lingual Speech-to-Intent Classification}
\name{Bidisha Sharma$^1$, Maulik Madhavi$^1$, Xuehao Zhou$^1$, Haizhou Li$^{1,2}$\thanks{This research is supported by the Agency for Science, Technology and Research (A*STAR) under its AME Programmatic Funding Scheme (Project No. A18A2b0046) and Science and Engineering Research Council, Agency of Science, Technology and Research, Singapore, through the National Robotics Program under Grant No. 192 25 00054. We thank the volunteers (native Mandarin speakers) for the manual correction of the Google translated Mandarin text from the English text.}}
\address{$^1$Department of Electrical and Computer Engineering,
National University of Singapore, Singapore\\
$^2$School of Data Science, The Chinese University of Hong Kong (Shenzhen), China
}
\begin{document}
%
\maketitle
\begin{abstract}
End-to-end speech-to-intent classification has shown its advantage in harvesting information from both text and speech. In this paper, we study a technique to develop such an end-to-end system that supports multiple languages. To overcome the scarcity of multi-lingual speech corpus, we exploit knowledge from a pre-trained multi-lingual natural language processing model. Multi-lingual bidirectional encoder representations from transformers (mBERT) models are trained on multiple languages and hence expected to perform well in the multi-lingual scenario. In this work, we employ a teacher-student learning approach to sufficiently extract information from an mBERT model to train a multi-lingual speech model. In particular, we use synthesized speech generated from an English-Mandarin text corpus for analysis and training of a multi-lingual intent classification model. We also demonstrate that the teacher-student learning approach obtains an improved performance (91.02\%) over the traditional end-to-end (89.40\%) intent classification approach in a practical multi-lingual scenario.
\end{abstract}

\vspace{-0.2cm}
\section{Introduction}
The rise in the application of voice-controlled devices and robots demands efficient intent classification from speech commands. A speech-based interface in the user's own language is highly desirable because it is a natural, intuitive, and flexible manner of communication~\cite{mulitlingual_intent_emnlp,castellucci2019multi}. In many societies, two or more languages are used in regular interaction in a spontaneous way~\cite{sharma2017polyglot,sharma2015development,mahanta2016text}. This prompts us to study multi-lingual spoken language understanding (SLU) techniques for human-machine or human-robot interaction.  

The recent studies show that end-to-end intent classification outperforms conventional pipeline systems that comprise of automatic speech recognition (ASR) and natural language understanding (NLU) modules~\cite{qian2017exploring,serdyuk2018towards}. The end-to-end intent classification can be broadly classified into two categories. 

The first category is the end-to-end modeling approach, where the features extracted from the speech signal are directly mapped to intents without utilising ASR resources~\cite{serdyuk2018towards,haghani2018audio}. The study presented in \cite{serdyuk2018towards} explores the scope of extending end-to-end ASR to support NLU component and perform domain/intent classification from speech. The research in \cite{haghani2018audio} discusses the sequence-to-sequence model to optimize jointly ASR and NLU as multi-task learning objective. 

The second category of study leverages information extracted from pre-trained acoustic models developed on large speech corpora or NLU models trained for related tasks. In this type of approach, the pre-trained model is used to extract or transfer knowledge and perform the downstream intent classification task. Such approaches address the issue of the unavailability of a large speech database for end-to-end SLU systems.

In~\cite{lugosch2019speech}, a pre-trained ASR model is employed to extract acoustic features from speech signals. The extracted features are considered to be more informative and closer to the phonetic content as compared to the conventional cepstral features extracted from a speech signal. Training an end-to-end network for intent classification using features extracted from the pre-trained ASR model is found to be effective~\cite{lugosch2019speech, radfar2020end,wang2020large,lugosch2019using,qian2021speech}. The study presented in \cite{wang2020large,radfar2020end} uses a transformer-based network to extract semantic context for intent classification. Another way to address the issue of limited data is to employ specific neural network architectures which perform well for a relatively small amount of data~\cite{lugosch2019speech, renkens2018capsule, poncelet2020multitask, mitra2019leveraging}. Using synthesized speech to increase SLU resources has been found to be effective in~\cite{lugosch2019using}.

Recently, the effectiveness of transfer learning from a pre-trained model has gained substantial attention in the research community across different areas~\cite{sharma2019automatic}. Such teacher-student based transfer learning methods aim to effectively infer the knowledge from a large teacher model to the student model for a specific task, which requires much less task-specific data than the conventional methods. The teacher-student based learning has also been explored in the cross-modal scenario, such as vision-language pre-training (VLP) model~\cite{zhou2020unified} for image captioning and SpeechBERT ~\cite{chuang2019speechbert} for SLU. 
A similar idea has been successfully adopted in intent classification task by aligning the acoustic embeddings extracted from an ASR system to the linguistic embeddings extracted from a language model, such as bidirectional encoder representations from transformers (BERT)~\cite{ denisov2020pretrained,sharma2021leveraging,jiang21c_interspeech}. In these studies, the BERT model acts as the teacher model and the SLU model acts as the student model for intent classification.

Despite much progress, SLU systems are not well explored in multi-lingual scenarios, whereas the practical applications demand adaptability of SLU for multiple languages. 
This raises several questions in one's thought, such as adaptability of one system developed for one language to another language, portability of an SLU system to adapt a new language, and training a single system for multiple languages. In this work, we aim to develop a single end-to-end intent classification system for multiple languages. We study a neural solution on two languages (English and Mandarin), which can be extended into multiple languages. Simply speaking, we build a single language-independent system to predict user's intent by taking either English or Mandarin speech as input.
\begin{figure}[t]
 \centering
\centerline{\epsfig{figure=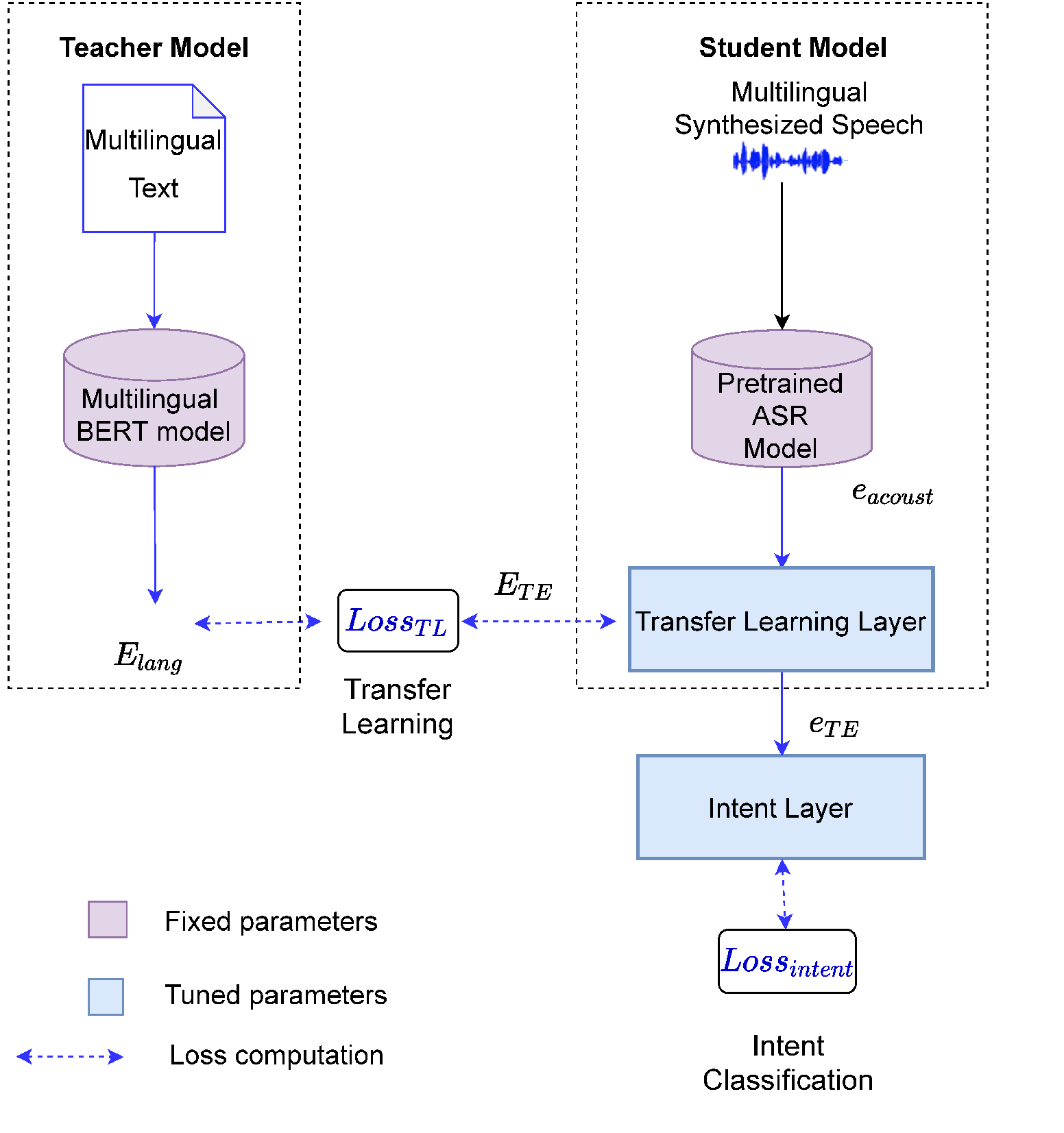,scale=0.50}}
\vspace{-.4cm}
{\ninept \caption{Block diagram of proposed multi-lingual teacher-student network (MTSN) for intent classification.}
\label{fig:block-diagram}}
\vspace{-0.4cm}
\end{figure}

In the same context, the multi-lingual BERT (mBERT) model has shown efficacy in terms of modeling multi-lingual intent detection and slot filling model for NLU~\cite{mulitlingual_intent_emnlp,castellucci2019multi}. To leverage the effectiveness of the mBERT model in the multi-lingual scenario, we propose a teacher-student learning approach, where the teacher model is an mBERT model and the student model is a neural network based model for intent classification. 

In this work, we would like to explore the efficacy of transfer learning to distill knowledge from the mBERT model and apply it effectively for end-to-end intent classification. To over the scarcity of multi-lingual speech data, we initially use a corpus of English-Mandarin synthesized speech in this study.  We also show the efficacy of the proposed method using a limited amount of recorded data. The proposed approach is resource effective from two aspects, firstly, the cross-domain transfer learning helps to accelerate inference from the mBERT model with restricted resources. Secondly, the use of synthesized speech for transfer learning demonstrates another step in developing speech based multi-lingual models in a resource efficient manner.





\vspace{-0.2cm}
\section{Multi-lingual Teacher-Student Network}\label{sec:MTSN}

In this section, we formulate a multi-lingual teacher-student network (MTSN), as shown in  Figure \ref{fig:block-diagram}. We adopt the mBERT model as the teacher model,  which is  pre-trained on a large volume database from 104 different languages with the largest Wikipedias with a shared vocabulary size of 110,000~\cite{devlin2019bert}.
  The student model contains a pre-trained ASR model, a transfer learning layer, and an intent layer. The transfer learning layer is responsible for leveraging knowledge from the mBERT model, which aims to minimize the distance between the linguistic embeddings obtained using the teacher model and acoustic embeddings obtained from the pre-trained ASR model. Next, the intent layer acts as the classification layer and outputs the predicted intent. 

\vspace{-0.2cm}
\subsection{Pre-trained ASR Model}
\label{sec:pretrained-nlu}
In this paper, we employ the same pre-trained ASR model used in ~\cite{lugosch2019speech}.
This model is a deep neural network consisting of a stack of modules. The first module (phoneme module) takes the audio signal as input and outputs a sequence of hidden representations to characterize the phonetic content of speech. This module uses a SincNet layer, which processes the raw input waveform, followed by multiple convolutional layers and recurrent layers with pooling (to reduce the sequence length) and dropout, which output a hidden phoneme representation ($h_{phoneme}$). These hidden phoneme representations are passed through a linear classifier and trained to predict phonemes.

The second module (word module) takes $h_{phoneme}$ obtained from the phoneme module as input and outputs 256-dimensional hidden word representation ($h_{word}$), which we refer to as acoustic embeddings  in this work. This module uses recurrent layers with dropout and pooling, and is pre-trained to predict whole word targets using another linear classifier. Further details of this pre-trained ASR model can be found in~\cite{lugosch2019speech}.  

This pre-trained ASR model is trained using a combination of English (Librispeech) and Mandarin (AISHELL) databases with separate phonesets. The network is trained to optimize the combined word and phoneme classification loss. After the ASR training is completed, we use this pre-trained ASR model by discarding the word and phoneme classifiers, and use the 256-dimensional hidden word representation ($h_{word}$) as acoustic embeddings ($e_{acoust}$).
We note that we freeze this pre-trained ASR model (i.e., fixed the parameters) during SLU training, and use it for the downstream intent classification task. 

\vspace{-0.3cm}
\subsection{Multi-lingual BERT Model}\label{Multilingual_BERT_Model}
One of the main issues in many NLP studies is the lack of a large amount of annotated data. 
The BERT model takes advantage of large unlabeled data during pre-training, and models better linguistic representations \cite{devlin2019bert}.  It can be further  fine-tuned for downstream tasks with a small amount of annotated data. 
The study presented in \cite{ACL_PiresSG19} shows that mBERT is effective in generalizing cross-lingual information for several tasks and able to perform zero-shot cross-lingual model transfer learning for SLU with limited available data. 

We apply the text for each utterance to the mBERT model and derive a sequence of 768-dimensional linguistic embeddings ($e_{lang}$). On the sequence output ($e_{lang}$), we apply mean-pooling to get sentence embedding, i.e., $E_{Lang}={MeanPool}(e_{lang})$. The model parameters used in the mBERT model are fixed during the training process. We note the mBERT model in this work can be replaced by any other state-of-the-art multi-lingual language model such as XLMR~\cite{conneau2019unsupervised}.

\vspace{-0.3cm}
\subsection{Transfer Learning Layer}\label{Transfer_Learning_Layer}
The transfer learning layer is responsible to perform knowledge distillation and learn the rich linguistic representation from mBERT, i.e., linguistic embedding $E_{lang}$. Here, 256-dimensional acoustic embeddings ($e_{acoust}$) are converted to a 768-dimensional representation using a linear transformation layer. We refer to this new derived representation as transferred embedding sequence (${e}_{TE}$). We then perform mean pooling over sequence length of ${e}_{TE}$ to derive transferred embedding (${E}_{TE}$). We follow the teacher-student learning method as presented in ~\cite{hinton2015distilling,li2017large} and the transfer learning layer performs knowledge distillation. The objective here is to make the transferred embeddings closer to the linguistic embeddings. To achieve this, we use the Kullback-Leibler (KL) divergence loss ($Loss_{TL}$) between the transferred embeddings and linguistic embeddings as discussed in~\cite{hinton2015distilling}. 

\vspace{-0.2cm}
\subsection{Intent Layer}\label{Intent_Classification_Layer}
The intent layer is responsible for predicting the intents from the transferred embedding sequence obtained as the output of the transfer learning layer. The first part of the intent layer module is a recurrent neural network (RNN) to characterize the temporal dynamics from the transferred embedding sequence, which is a gated recurrent unit (GRU). The output from the GRU is fed to max-pooling and linear transformation layers for intent classification. The intent classification loss is denoted by ${Loss}_{intent}$ and computed using cross-entropy calculation as follows.
\begin{equation}\label{Intent_loss}
Loss_{intent}=\text{CrossEntropy}(g(e_{TL}),y),
\vspace{-0.1cm}
\end{equation}
where, $g(e_{TL})$ is the intent layer output, $e_{TL}$ is the input to intent layer and $y$ represents intent labels. We combine the two losses, namely, intent loss (${Loss}_{intent}$) and the transfer learning loss (${Loss}_{TL}$) using the interpolation weight $\alpha$ to derive the total loss ($Loss_{total}$), which is used to back-propagate the MTSN framework.
\begin{equation}
    Loss_{{total}} = \alpha Loss_{TL} + (1- \alpha) Loss_{intent}
    \label{eq:totalloss}
    \vspace{-0.1cm}
\end{equation}

\vspace{-0.2 cm}
\section{Experiments}
\label{sec:experiments}
In this section, we discuss the database and experimental details for the MTSN experiments. We also introduce the baseline systems for comparative studies.  

\vspace{-0.3cm}
\subsection{Database}
\label{DatabaseSection}
In this work, we consider two languages, namely, English and Mandarin to support our idea for multi-lingual intent classification. As developing a new sufficiently large training corpus for the SLU system with parallel multi-lingual utterances is challenging, initially we consider synthesized speech obtained from a multi-lingual text-to-speech (TTS) system. We also develop a relatively smaller recorded database for evaluation in practical multi-lingual scenario.

We use the utterances from the Fluent speech commands (FSC) dataset~\cite{lugosch2019speech} as base text material for designing of multi-lingual synthesized and recorded speech corpora. The FSC database consists of 248 unique utterances with 31 unique intent classes, which serve as speech commands to a virtual assistant. To generate the Mandarin prompts from English text, we conduct a two-stage translation approach. In the first stage, we translate these utterances to Mandarin using Google's English to Mandarin translator\footnote{https://translate.google.com/}. In the second stage, we conduct further manual correction of translated utterances by native Mandarin speakers.

\vspace{-0.2cm}
\subsubsection{Synthesized English-Mandarin corpus}
\label{sec:TTS}
Both the English and Mandarin spoken utterances are synthesized from a pre-trained multi-speaker multi-lingual TTS system.
The TTS system is trained based on the modified Tacotron2 \cite{shen2018natural} architecture, a sequence-to-sequence network predicting Mel-spectrogram from the text directly. We use the phone as input representation and concatenate language embedding with phone embedding. The residual encoder structure \cite{xue2019building} is adopted, and we combine the pre-trained cross-lingual word embedding with the residual encoder output to increase the cross-lingual linguistic information sharing, as in~\cite{zhou2020end}. To control the speaker identity, we use the speaker look-up table. The speaker embedding is combined with decoder long-short term memory (LSTM) input as in \cite{xue2019building}. The speech audios are synthesized from Mel-spectrogram using Griffin-Lim \cite{griffin1984signal} algorithm. The mean squared error (MSE) is used for frame reconstruction and guided-attention loss \cite{tachibana2018efficiently} is adopted to speed up the alignment process.
The TTS system is trained using a Mandarin monolingual dataset and an English monolingual dataset. We use a total of 11,098 Mandarin utterances from 52 Mandarin speakers from THCHS30 \cite{wang2015thchs}  and a total of 10,914 English utterances from 110 English speakers from LibriTTS  \cite{zen2019libritts}, and one bilingual speaker \footnote{https://www.data-baker.com/us\_en.html} with 200 Mandarin utterances and 150 English utterances. All audio samples are down-sampled to 16kHz sample rate.

From this TTS system, we choose 50 seen Mandarin monolingual speakers and let each of them synthesize the 248 Mandarin utterances. Similarly, we choose 50 seen English monolingual speakers and let each of them synthesize the 248 English utterances. In this way, we derive a multi-lingual English-Mandarin synthesized speech corpus with 12,800 utterances and corresponding intent labels for each of the English and Mandarin languages. The dataset is divided into train and test sets so that there is no overlap of utterances between the two. In the original FSC dataset, the train, test, and dev sets contain similar text prompts from different speakers. We believe that the use of similar text in train and test might not be sensible for practical purposes. Further details of this synthesized English-Mandarin speech corpus for multi-lingual intent classification are presented in Table~\ref{Database}.

\begin{table}[t]
\vspace{0.3cm} 
\caption{Synthesized English-Mandarin speech corpus.}
\label{Database} 
\renewcommand{\arraystretch}{1.2}
\resizebox{0.45\textwidth}{!}{
\begin{tabular}{|c c c c c c c|}
\hline
\multicolumn{1}{|c|}{\multirow{1}{*}{Language$\rightarrow$}} & \multicolumn{2}{|c|}{\bf{Eng}}  & \multicolumn{2}{|c|}{\bf{Man}} & \multicolumn{2}{|c|}{\bf{Eng-Man}}\\
\hline
\multicolumn{1}{|c|}{Specification$\downarrow$} & \multicolumn{1}{|c|}{Train} & \multicolumn{1}{|c|}{Test} & \multicolumn{1}{|c|}{Train} & \multicolumn{1}{|c|}{Test} & \multicolumn{1}{|c|}{Train} & \multicolumn{1}{|c|}{Test}\\
\hline
\hline
\multicolumn{1}{|c|}{\#Utterances} & \multicolumn{1}{|c|}{9900} & \multicolumn{1}{|c|}{2500} & \multicolumn{1}{|c|}{9900} & \multicolumn{1}{|c|}{2500} & \multicolumn{1}{|c|}{19,800} & \multicolumn{1}{|c|}{5000}\\
\hline
\multicolumn{1}{|c|}{\#Unique utterances} & \multicolumn{1}{|c|}{198} & \multicolumn{1}{|c|}{50} & \multicolumn{1}{|c|}{198} & \multicolumn{1}{|c|}{50} & \multicolumn{1}{|c|}{396} & \multicolumn{1}{|c|}{100}\\
\hline
\multicolumn{1}{|c|}{Duration (min)} & \multicolumn{1}{|c|}{244.21} & \multicolumn{1}{|c|}{61.75} & \multicolumn{1}{|c|}{277.25} & \multicolumn{1}{|c|}{71.13} & \multicolumn{1}{|c|}{521.46} & \multicolumn{1}{|c|}{132.88}\\
\hline
\multicolumn{1}{|c|}{\#Speakers} & \multicolumn{1}{|c|}{50} & \multicolumn{1}{|c|}{50} & \multicolumn{1}{|c|}{50} & \multicolumn{1}{|c|}{50} & \multicolumn{1}{|c|}{100} & \multicolumn{1}{|c|}{100}\\
\hline
\end{tabular}
}
\vspace{-.3cm}
\end{table}

\begin{figure*}[t!]
\centering
\centerline{\epsfig{figure=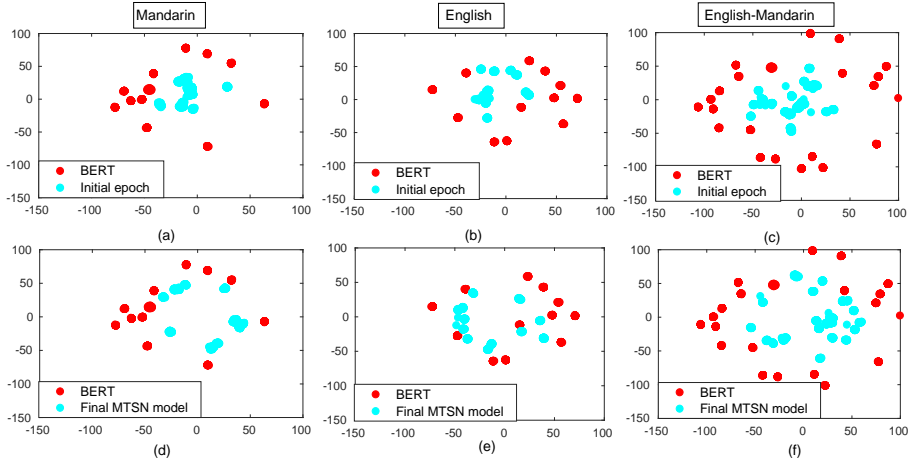, height=2.4in,width=4.8in}}
\vspace{-.2cm}
{\ninept\caption{t-SNE visualization plots to compare embeddings from initial epoch and final MTSN model againt mBERT embeddings (a) initial epoch, (d) final MTSN model for Mandarin test data; (b) initial epoch, (e) final MTSN model for English test data; (c) initial epoch, (f) final MTSN model for English-Mandarin test data.}\label{fig:TSNE}}
\end{figure*}

\vspace{-.2cm}
\subsubsection{Recorded English-Mandarin corpus}\label{sec:Recorded}
We record the same 248 utterances of each English and Mandarin text from 11 native Mandarin speakers who are fluent in both the languages. We manually verify the recorded data resulting in 2,695 English and 2,708 Mandarin utterances. We divide this data into train and test sets following the same strategy as that of the synthesized speech corpus. The quantitative detail of the recorded English-Mandarin corpus is depicted in Table~\ref{DatabaseRecorded}.

\begin{table}[t]
\vspace{0.3cm} 
\caption{Recorded English-Mandarin speech corpus.}
\label{DatabaseRecorded} 
\renewcommand{\arraystretch}{1.2}
\resizebox{0.45\textwidth}{!}{
\begin{tabular}{|c c c c c c c|}
\hline
\multicolumn{1}{|c|}{\multirow{1}{*}{Language$\rightarrow$}} & \multicolumn{2}{|c|}{\bf{Eng}}  & \multicolumn{2}{|c|}{\bf{Man}} & \multicolumn{2}{|c|}{\bf{Eng-Man}}\\
\hline
\multicolumn{1}{|c|}{Specification$\downarrow$} & \multicolumn{1}{|c|}{Train} & \multicolumn{1}{|c|}{Test} & \multicolumn{1}{|c|}{Train} & \multicolumn{1}{|c|}{Test} & \multicolumn{1}{|c|}{Train} & \multicolumn{1}{|c|}{Test}\\
\hline
\hline
\multicolumn{1}{|c|}{\#Utterances} & \multicolumn{1}{|c|}{2205} & \multicolumn{1}{|c|}{490} & \multicolumn{1}{|c|}{2218} & \multicolumn{1}{|c|}{490} & \multicolumn{1}{|c|}{4417} & \multicolumn{1}{|c|}{980}\\
\hline
\multicolumn{1}{|c|}{\#Unique utterances} & \multicolumn{1}{|c|}{198} & \multicolumn{1}{|c|}{50} & \multicolumn{1}{|c|}{198} & \multicolumn{1}{|c|}{50} & \multicolumn{1}{|c|}{396} & \multicolumn{1}{|c|}{100}\\
\hline
\multicolumn{1}{|c|}{Duration (min)} & \multicolumn{1}{|c|}{87.41} & \multicolumn{1}{|c|}{23.47} & \multicolumn{1}{|c|}{96.78} & \multicolumn{1}{|c|}{20.82} & \multicolumn{1}{|c|}{183.95} & \multicolumn{1}{|c|}{44.30}\\
\hline
\multicolumn{1}{|c|}{\#Speakers} & \multicolumn{1}{|c|}{9} & \multicolumn{1}{|c|}{2} & \multicolumn{1}{|c|}{9} & \multicolumn{1}{|c|}{2} & \multicolumn{1}{|c|}{9} & \multicolumn{1}{|c|}{2}\\
\hline
\end{tabular}
}
\end{table}

\vspace{-.2cm}
\subsection{Experimental Setup}\label{Setup}
\subsubsection{Baseline systems}\label{Baseline}
To benchmark the proposed MTSN method for multi-lingual intent classification, we develop two baseline systems. 

First, {\bf Baseline-1} is the traditional pipeline approach, where we derive text from English and Mandarin ASR systems, which is then taken by the mBERT model for intent classification. The ASR systems employed here are in-house general purpose English and Mandarin ASR models, trained  using  a  combination of several  speech  databases for English and Mandarin languages, respectively. The second baseline ({\bf Baseline-2}) is deployed using acoustic features extracted from the pre-trained ASR model~\cite{lugosch2019speech}. In {\bf Baseline-2}, we first apply the spoken utterances to the English-Mandarin pre-trained ASR model as discussed in Section~\ref{sec:pretrained-nlu}~\footnote{https://github.com/lorenlugosch/end-to-end-SLU} and derive the 256-dimensional acoustic embeddings. We apply the acoustic embeddings to the intent layer as discussed in Section~\ref{Intent_Classification_Layer} and back-propagate the system to minimize the intent loss ($Loss_{intent}$). We note that we do not use the transfer learning loss ($Loss_{TE}$) while training the Baseline-2 system. This framework does not learn from linguistic knowledge of mBERT. We train 3 language specific Baseline-2 systems using English, Mandarin, and both as train dataset. 

\subsubsection{Propsoed MTSN system}\label{Setup}

To develop the multi-lingual intent classification system using the proposed MTSN framework ({\bf Proposed MTSN}), we first pass the English-Mandarin spoken utterances through the pre-trained ASR model to obtain the 256-dimensional acoustic embeddings as in Section \ref{sec:pretrained-nlu}.
Similarly, we input the corresponding text to the mBERT model to obtain 768-dimensional linguistic embedding as discussed in Section~\ref{Multilingual_BERT_Model}. Next, using the transfer learning layer as in Section~\ref{Transfer_Learning_Layer} we obtain the 768-dimensional transferred embedding sequence ($e_{TE}$). We believe that the transferred embedding sequences are generated by taking knowledge from the mBERT teacher model and hence expected to capture better linguistic information. We apply these transferred embedding sequences to the intent layer as discussed in Section~\ref{Intent_Classification_Layer}. The entire framework is trained considering the combination of $Loss_{TL}$ and $Loss_{intent}$ as in Equation~\ref{eq:totalloss}.  We train 3 language specific Proposed MTSN systems using English, Mandarin, and both as train dataset. We use $\alpha=0.5$ to give equal emphasis to both the losses while calculating $Loss_{total}$.

All the models (baseline and proposed) are deployed using PyTorch. We use the Adam optimizer with a learning rate of 0.001, batch size 32, and 100 epochs.

\vspace{-0.2cm}
\section{Experimental Results}\label{sec:results}
\vspace{-0.2cm}
\subsection{Visualization of Embeddings}
To demonstrate the effectiveness of the proposed teacher-student model based transfer learning approach in a multi-lingual scenario, we first analyse 768-dimensional transfer embeddings ($E_{TE}$) obtained at an initial epoch and after training the model, and compare against embeddings extracted from the mBERT model ($E_{lang}$). We visualize t-distributed stochastic neighbor embedding (t-SNE) representation~\cite{maaten2008visualizing} of these embeddings. 

In Figure~\ref{fig:TSNE}, we show t-SNE representation of transfer embeddings at an initial epoch and final MTSN model, while testing the model with Mandarin, English, and English-Mandarin data for 10 intent classes. We can observe that in Figure~\ref{fig:TSNE} (d), (e) and (f) the distribution of the transfer embedding ($E_{TE}$) clusters are closer to the mBERT embeddings ($E_{lang}$) than those of Figure~\ref{fig:TSNE} (a), (b) and (c). We further observe that for the English-Mandarin case, there is a higher number of clusters representing both the  languages. To quantify this evidence, we calculate the cosine similarity of transfer embeddings obtained from the initial epoch and proposed model with mBERT embeddings for English, Mandarin, and English-Mandarin test data. We calculate average cosine similarity over all utterances of the test datasets. We observe that the embeddings after transfer learning are closer to BERT embeddings than the initial epoch embeddings as in Table~\ref{Cosine}. We use the recorded English-Mandarin speech data for this analysis.

\begin{table}[!t]
\centering
\caption{Average cosine similarity between transferred embeddings obtained from initial epoch, final model with embeddings obtained from mBERT for English, Mandarin and English-Mandarin test data.}
\label{Cosine} 
\vspace{0.5cm}
\renewcommand{\arraystretch}{1}
\resizebox{0.48\textwidth}{!}{
\centerline{
\begin{tabular}{|c c c|}
\hline
\multicolumn{1}{|c|}{\bf Test language}
& \multicolumn{1}{|c|}{\bf{Initial epoch}}  & \multicolumn{1}{|c|}{\bf{Final MTSN model}}\\
\hline\hline
\multicolumn{1}{|c|}{Eng} & \multicolumn{1}{|c|}{0.0156} & \multicolumn{1}{|c|}{0.0362}\\
\hline
\multicolumn{1}{|c|}{Man} & \multicolumn{1}{|c|}{-0.0107} & \multicolumn{1}{|c|}{0.0307}\\
\hline
\multicolumn{1}{|c|}{Eng-Man} & \multicolumn{1}{|c|}{0.0024} & \multicolumn{1}{|c|}{0.0335}\\
\hline
\end{tabular}
}}
\end{table}
\subsection{Experiments on Synthesized Corpus}
In Table \ref{tab:proposed-result}, we show the experimental results corresponding to the synthesized English-Mandarin speech corpus. We obtain the performance for different language specific train and test datasets for Baseline-2 and Proposed MTSN models. For Baseline-1, we use the pre-trained mBERT model and test for different languages.  It can be observed from the Table \ref{tab:proposed-result} that the Baseline-2 system gives better performance in language matching scenario, which is intuitive. For example, the Baseline-2 system trained with English gives 85.20 \% on English test set and gives a poor performance of 7.88 \% on Mandarin test set due to language mismatch. A similar trend in the result is observed for the Baseline-2 model trained with Mandarin language speech data.
 
Next, to eliminate the language mismatch effect in the baseline frameworks, we further use synthesized English-Mandarin training data. We observe that the performance is improved while using both the languages in training as compared to the language mismatched scenarios. The Proposed MTSN framework exploits the knowledge from the linguistically rich mBERT model and is expected to perform better for intent classification. We use the same synthesized monolingual and English-Mandarin training data to develop the proposed MTSN model. We observe that the monolingual intent classification performance is also improved using the proposed MTSN model compared to that of Baseline-2. However, as expected the performance of monolingual intent classification using the proposed method is superior to that of multi-lingual intent classification. For all the systems, the intent classification accuracy is better using the proposed MTSN model than both Baseline-1 and Baseline-2 systems.

\begin{table}[!t]
\centering
\caption{Intent classification performance with proposed MTSN and baseline frameworks w.r.t. different language scenarios using synthesized English-Mandarin corpus.}
\vspace{0.6cm}
\label{tab:proposed-result}
\scalebox{0.85}{
\begin{tabular}{|c |c |c|c|}
\hline
{\multirow{2}{*}{\textbf{Framework}}} & \textbf{Train}  & \textbf{Test}   & \textbf{Accuracy} \\ 
 & \textbf{language}  & \textbf{language}   & \textbf{(\%)}\\\hline\hline
{\multirow{3}{*}{Baseline-1}}      & {\multirow{3}{*}{mBERT model}}         & Eng         & 80.78     \\\cline{3-4}
       &          & Man         & 87.88     \\\cline{3-4}
       &         & Eng-Man         & 84.33     
    \\\hline\hline
{\multirow{7}{*}{Baseline-2}}      & {\multirow{2}{*}{Eng}}         & Eng         & 85.20     \\\cline{3-4}
       &          & Man         & 7.88     \\\cline{2-4}
       & {\multirow{2}{*}{Man}}         & Eng         & 9.28     \\\cline{3-4}
       &         & Man         & 93.32    \\\cline{2-4}
       & {\multirow{3}{*}{Eng-Man}}     & Eng        & 75.76    \\\cline{3-4}
            &  &  Man         & 92.52    \\\cline{3-4}
            &   & Eng-Man & 84.14    \\\hline\hline
{\multirow{7}{*}{Proposed MTSN}}      & {\multirow{2}{*}{Eng}}         & Eng         & 89.30     \\\cline{3-4}
       &          & Man         & 6.84     \\\cline{2-4}
       & {\multirow{2}{*}{Man}}         & Eng         & 8.50     \\\cline{3-4}
       &         & Man         & 96.43    \\\cline{2-4}
        & {\multirow{3}{*}{Eng-Man}}      & Eng         & \textbf{81.64} \\\cline{3-4}
       &      & Man         & \textbf{93.48}    \\\cline{3-4}
       &      & Eng-Man & \textbf{87.56}   \\\hline
\end{tabular}}
\end{table}
\subsubsection{Experiments on limited training data scenario}
Further, to observe the efficacy of the proposed MTSN model in case of less training data, we consider different subsets of the synthesized English-Mandarin training corpus. In this case, we choose 10\%,  50\% and  70\% of the entire training data to develop different versions of the English-Mandarin baseline and proposed models. From Table~\ref{Data_partition} we can observe a very small change in the intent classification accuracy with the reduction of training data for the proposed MTSN model compared to that of Baseline-1 and Baseline-2.

\begin{table}[!t]
\centering
\caption{Intent classification performance with proposed MTSN and baseline frameworks w.r.t. different subsets of training data using synthesized English-Mandarin corpus.}
\label{Data_partition} 
\vspace{0.5cm}
\renewcommand{\arraystretch}{1}
\resizebox{0.45\textwidth}{!}{
\centerline{
\begin{tabular}{|c c c c|}
\hline
\multicolumn{1}{|c|}{\multirow{2}{*}{\backslashbox{\bf Framework}{\bf Training data}}}
& \multicolumn{3}{|c|}{\bf{Accuracy(\%)}}\\
\cline{2-4}
\multicolumn{1}{|c|}{}
& \multicolumn{1}{|c|}{\bf{10\%}}  & \multicolumn{1}{|c|}{\bf{50\%}}  & \multicolumn{1}{|c|}{\bf{70\%}}\\
\hline\hline
\multicolumn{1}{|c|}{Baseline-1} & \multicolumn{1}{|c|}{78.30} & \multicolumn{1}{|c|}{80.55} & \multicolumn{1}{|c|}{83.50}\\
\hline
\multicolumn{1}{|c|}{Baseline-2}  & \multicolumn{1}{|c|}{80.64} & \multicolumn{1}{|c|}{82.18} & \multicolumn{1}{|c|}{83.36}\\
\hline
\multicolumn{1}{|c|}{Proposed MTSN} & \multicolumn{1}{|c|}{86.10} & \multicolumn{1}{|c|}{87.02} & \multicolumn{1}{|c|}{87.34}\\
\hline
\end{tabular}
}}
\end{table}
\subsection{Experiments on Recorded Corpus}
The above mentioned experiments are performed on synthesized database which is comparatively larger than our recorded multi-lingual database. To demonstrate the efficacy of the proposed method in realistic scenario we develop the Baseline-1, Baseline-2 and Proposed MTSN models using the recorded English-Mandarin corpus discussed in Section~\ref{sec:Recorded}. In Table~\ref{Result_recorded}, we show the performance of the baseline and proposed models for the recorded data. Similar to the synthesized speech results we observe the intent classification accuracy of the proposed MTSN model for English-Mandarin data is 91.02\%, which is 88.64\% for the Baseline-1 and 89.40\% for Baseline-2 systems. This demonstrates the efficacy of the teacher-student approach to learn embeddings from a rich linguistic resource, and impose on task specific SLU models in the multi-lingual scenario.

From the results depicted in Table~\ref{tab:proposed-result} and Table~\ref{Result_recorded}, we observe that for both synthesized and recorded data the performance of Mandarin is superior to that of English, which may be because of the simpler grammatical structure of the sentences in Mandarin compared to that of English. This creates less confusion among similar utterances with different intents in case of Mandarin language. Apart of from the multi-lingual scenario we also observe an improvement in performance of the proposed approach in the monolingual case. This eventually is a result of the transfer learning method. We also note that the performance of multi-lingual model is not as good as the monolingual counterpart.

\begin{table}[!t]
\centering
\caption{ Intent classification performance with proposed MTSN and baseline frameworks trained using recorded English-Mandarin corpus for different test languages.}
\label{Result_recorded} 
\vspace{0.3cm}
\renewcommand{\arraystretch}{1.2}
\resizebox{0.45\textwidth}{!}{
\centerline{
\begin{tabular}{|c c c c|}
\hline
\multicolumn{1}{|c|}{\multirow{2}{*}{\backslashbox{\bf Framework}{\bf Training data}}}
& \multicolumn{3}{|c|}{\bf{Accuracy(\%)}}\\
\cline{2-4}
\multicolumn{1}{|c|}{}
& \multicolumn{1}{|c|}{\bf{Eng}}  & \multicolumn{1}{|c|}{\bf{Man}}  & \multicolumn{1}{|c|}{\bf{Eng-Man}}\\
\hline\hline
\multicolumn{1}{|c|}{Baseline-1} & \multicolumn{1}{|c|}{84.78} & \multicolumn{1}{|c|}{92.50} & \multicolumn{1}{|c|}{88.64}\\
\hline
\multicolumn{1}{|c|}{Baseline-2}  & \multicolumn{1}{|c|}{86.50} & \multicolumn{1}{|c|}{92.30} & \multicolumn{1}{|c|}{89.40}\\
\hline
\multicolumn{1}{|c|}{Proposed MTSN} & \multicolumn{1}{|c|}{87.55} & \multicolumn{1}{|c|}{93.46} & \multicolumn{1}{|c|}{91.02}\\
\hline
\end{tabular}
}}
\end{table}

\section{Conclusions}
\label{sec:conclusion}
In this work, we explore the teacher-student learning approach to transfer the knowledge from a linguistic model to a speech based end-to-end intent classification model, for multi-lingual spoken utterances. The proposed MTSN framework leverages knowledge distillation from an mBERT model to the intent classification model. The mBERT model can be replaced by other state-of-the-art language models~\cite{conneau2019unsupervised}. Although the aim of this work is to explore the capability of transfer learning in multi-lingual scenario, currently we consider only two languages, English and Mandarin. In the future we aim to extend this work further by incorporating more languages. For evaluation of the proposed idea, we use English-Mandarin synthesized speech corpus along with a relatively smaller recorded speech corpus. The experimental evidence shows that the proposed model is able to give better performance compared to the conventional end-to-end intent classification model for multi-lingual data.  


\newpage
\bibliographystyle{IEEEbib}
\balance
\bibliography{Multilingual_intent.bib}

\begin{thebibliography}{10}

\bibitem{mulitlingual_intent_emnlp}
Talaat Khalil, Kornel Kielczewski, Georgios~Christos Chouliaras, Amina
  Keldibek, and Maarten Versteegh,
\newblock ``Cross-lingual intent classification in a low resource industrial
  setting,''
\newblock in {\em Proc. {EMNLP-IJCNLP}}. 2019, pp. 6418--6423, Association for
  Computational Linguistics.

\bibitem{castellucci2019multi}
Giuseppe Castellucci, Valentina Bellomaria, Andrea Favalli, and Raniero
  Romagnoli,
\newblock ``Multi-lingual intent detection and slot filling in a joint
  bert-based model,''
\newblock {\em arXiv preprint arXiv:1907.02884}, 2019.

\bibitem{sharma2017polyglot}
Bidisha Sharma and SR~Mahadeva Prasanna,
\newblock ``Polyglot speech synthesis: a review,''
\newblock {\em IETE Technical Review}, vol. 34, no. 4, pp. 366--389, 2017.

\bibitem{sharma2015development}
Bidisha Sharma, Nagaraj Adiga, and SR~Mahadeva Prasanna,
\newblock ``Development of assamese text-to-speech synthesis system,''
\newblock in {\em TENCON 2015-2015 IEEE Region 10 Conference}. IEEE, 2015, pp.
  1--6.

\bibitem{mahanta2016text}
Deepshikha Mahanta, Bidisha Sharma, Priyankoo Sarmah, and SR~Mahadeva Prasanna,
\newblock ``Text to speech synthesis system in indian english,''
\newblock in {\em 2016 IEEE Region 10 Conference (TENCON)}. IEEE, 2016, pp.
  2614--2618.

\bibitem{qian2017exploring}
Yao Qian, Rutuja Ubale, Vikram Ramanaryanan, Patrick Lange, David
  Suendermann-Oeft, Keelan Evanini, and Eugene Tsuprun,
\newblock ``Exploring {ASR}-free end-to-end modeling to improve spoken language
  understanding in a cloud-based dialog system,''
\newblock in {\em Automatic Speech Recognition and Understanding Workshop
  (ASRU)}. 2017, pp. 569--576, {IEEE}.

\bibitem{serdyuk2018towards}
Dmitriy Serdyuk, Yongqiang Wang, Christian Fuegen, Anuj Kumar, Baiyang Liu, and
  Yoshua Bengio,
\newblock ``Towards end-to-end spoken language understanding,''
\newblock in {\em International Conference on Acoustics, Speech and Signal
  Processing (ICASSP)}. IEEE, 2018, pp. 5754--5758.

\bibitem{haghani2018audio}
Parisa Haghani, Arun Narayanan, Michiel Bacchiani, Galen Chuang, Neeraj Gaur,
  Pedro Moreno, Rohit Prabhavalkar, Zhongdi Qu, and Austin Waters,
\newblock ``From audio to semantics: Approaches to end-to-end spoken language
  understanding,''
\newblock in {\em IEEE Spoken Language Technology Workshop (SLT)}, 2018, pp.
  720--726.

\bibitem{lugosch2019speech}
Loren Lugosch, Mirco Ravanelli, Patrick Ignoto, Vikrant~Singh Tomar, and Yoshua
  Bengio,
\newblock ``Speech model pre-training for end-to-end spoken language
  understanding,''
\newblock in {\em Proc. Interspeech}, 2019, pp. 814--818.

\bibitem{radfar2020end}
Martin Radfar, Athanasios Mouchtaris, and Siegfried Kunzmann,
\newblock ``{End-to-End Neural Transformer Based Spoken Language
  Understanding},''
\newblock in {\em Proc. Interspeech}, 2020, pp. 866--870.

\bibitem{wang2020large}
Pengwei Wang, Liangchen Wei, Yong Cao, Jinghui Xie, and Zaiqing Nie,
\newblock ``Large-scale unsupervised pre-training for end-to-end spoken
  language understanding,''
\newblock in {\em International Conference on Acoustics, Speech and Signal
  Processing (ICASSP)}. IEEE, 2020, pp. 7999--8003.

\bibitem{lugosch2019using}
Loren Lugosch, Brett Meyer, Derek Nowrouzezahrai, and Mirco Ravanelli,
\newblock ``Using speech synthesis to train end-to-end spoken language
  understanding models,''
\newblock in {\em International Conference on Acoustics, Speech and Signal
  Processing (ICASSP)}. IEEE, 2020, pp. 8499--8503.

\bibitem{qian2021speech}
Yao Qian, Ximo Bian, Yu~Shi, Naoyuki Kanda, Leo Shen, Zhen Xiao, and Michael
  Zeng,
\newblock ``Speech-language pre-training for end-to-end spoken language
  understanding,''
\newblock {\em arXiv preprint arXiv:2102.06283}, 2021.

\bibitem{renkens2018capsule}
Vincent Renkens and Hugo~Van hamme,
\newblock ``Capsule networks for low resource spoken language understanding,''
\newblock in {\em Proc. Interspeech}, 2018, pp. 601--605.

\bibitem{poncelet2020multitask}
Jakob Poncelet and Hugo~Van hamme,
\newblock ``Multitask learning with capsule networks for speech-to-intent
  applications,''
\newblock in {\em International Conference on Acoustics, Speech and Signal
  Processing (ICASSP)}. IEEE, 2020, pp. 8494--8498.

\bibitem{mitra2019leveraging}
Vikramjit Mitra, Sue Booker, Erik Marchi, David~Scott Farrar, Ute~Dorothea
  Peitz, Bridget Cheng, Ermine Teves, Anuj Mehta, and Devang Naik,
\newblock ``Leveraging acoustic cues and paralinguistic embeddings to detect
  expression from voice,''
\newblock in {\em Proc. Interspeech}, 2019, pp. 1651--1655.

\bibitem{sharma2019automatic}
Bidisha Sharma, Chitralekha Gupta, Haizhou Li, and Ye~Wang,
\newblock ``Automatic lyrics-to-audio alignment on polyphonic music using
  singing-adapted acoustic models,''
\newblock in {\em ICASSP 2019-2019 IEEE International Conference on Acoustics,
  Speech and Signal Processing (ICASSP)}. IEEE, 2019, pp. 396--400.

\bibitem{zhou2020unified}
Luowei Zhou, Hamid Palangi, Lei Zhang, Houdong Hu, Jason Corso, and Jianfeng
  Gao,
\newblock ``Unified vision-language pre-training for image captioning and
  {VQA},''
\newblock in {\em Proceedings of the AAAI Conference on Artificial
  Intelligence}, 2020, vol.~34, pp. 13041--13049.

\bibitem{chuang2019speechbert}
Yung-Sung Chuang, Chi-Liang Liu, Hung yi~Lee, and Lin shan Lee,
\newblock ``{SpeechBERT: An Audio-and-Text Jointly Learned Language Model for
  End-to-End Spoken Question Answering},''
\newblock in {\em Proc. Interspeech}, 2020, pp. 4168--4172.

\bibitem{denisov2020pretrained}
Pavel Denisov and Ngoc~Thang Vu,
\newblock ``{Pretrained Semantic Speech Embeddings for End-to-End Spoken
  Language Understanding via Cross-Modal Teacher-Student Learning},''
\newblock in {\em Proc. Interspeech}, 2020, pp. 881--885.

\bibitem{sharma2021leveraging}
Bidisha Sharma, Maulik Madhavi, and Haizhou Li,
\newblock ``Leveraging acoustic and linguistic embeddings from pretrained
  speech and language models for intent classification,''
\newblock in {\em ICASSP 2021-2021 IEEE International Conference on Acoustics,
  Speech and Signal Processing (ICASSP)}. IEEE, 2021, pp. 7498--7502.

\bibitem{jiang21c_interspeech}
Yidi Jiang, Bidisha Sharma, Maulik Madhavi, and Haizhou Li,
\newblock ``{Knowledge Distillation from BERT Transformer to Speech Transformer
  for Intent Classification},''
\newblock in {\em Proc. Interspeech 2021}, 2021, pp. 4713--4717.

\bibitem{devlin2019bert}
Jacob Devlin, Ming-Wei Chang, Kenton Lee, and Kristina Toutanova,
\newblock ``{BERT:} pre-training of deep bidirectional transformers for
  language understanding,''
\newblock in {\em North American Chapter of the Association for Computational
  Linguistics: Human Language Technologies, {NAACL-HLT}}, 2019, p. 4171–4186.

\bibitem{ACL_PiresSG19}
Telmo Pires, Eva Schlinger, and Dan Garrette,
\newblock ``How multilingual is multilingual bert?,''
\newblock in {\em Proceedings of the 57th Conference of the Association for
  Computational Linguistics, {ACL} 2019, Florence, Italy, July 28- August 2,
  2019, Volume 1: Long Papers}. 2019, pp. 4996--5001, Association for
  Computational Linguistics.

\bibitem{conneau2019unsupervised}
Alexis Conneau, Kartikay Khandelwal, Naman Goyal, Vishrav Chaudhary, Guillaume
  Wenzek, Francisco Guzm{\'a}n, Edouard Grave, Myle Ott, Luke Zettlemoyer, and
  Veselin Stoyanov,
\newblock ``Unsupervised cross-lingual representation learning at scale,''
\newblock {\em arXiv preprint arXiv:1911.02116}, 2019.

\bibitem{hinton2015distilling}
Geoffrey Hinton, Oriol Vinyals, and Jeffrey Dean,
\newblock ``Distilling the knowledge in a neural network,''
\newblock in {\em NIPS Deep Learning and Representation Learning Workshop},
  2015.

\bibitem{li2017large}
Jinyu Li, Michael~L Seltzer, Xi~Wang, Rui Zhao, and Yifan Gong,
\newblock ``Large-scale domain adaptation via teacher-student learning,''
\newblock in {\em Proc. Interspeech}, 2017, pp. 2386--2390.

\bibitem{shen2018natural}
Jonathan Shen, Ruoming Pang, Ron~J Weiss, Mike Schuster, Navdeep Jaitly,
  Zongheng Yang, Zhifeng Chen, Yu~Zhang, Yuxuan Wang, Rj~Skerrv-Ryan, et~al.,
\newblock ``Natural tts synthesis by conditioning wavenet on mel spectrogram
  predictions,''
\newblock in {\em 2018 IEEE International Conference on Acoustics, Speech and
  Signal Processing (ICASSP)}. IEEE, 2018, pp. 4779--4783.

\bibitem{xue2019building}
Liumeng Xue, Wei Song, Guanghui Xu, Lei Xie, and Zhizheng Wu,
\newblock ``Building a mixed-lingual neural tts system with only monolingual
  data.,''
\newblock in {\em Proc Interspeech}, 2019, pp. 2060--2064.

\bibitem{zhou2020end}
Xuehao Zhou, Xiaohai Tian, Grandee Lee, Rohan~Kumar Das, and Haizhou Li,
\newblock ``End-to-end code-switching tts with cross-lingual language model,''
\newblock in {\em International Conference on Acoustics, Speech and Signal
  Processing (ICASSP)}. IEEE, 2020, pp. 7614--7618.

\bibitem{griffin1984signal}
Daniel Griffin and Jae Lim,
\newblock ``Signal estimation from modified short-time fourier transform,''
\newblock {\em IEEE Transactions on acoustics, speech, and signal processing},
  vol. 32, no. 2, pp. 236--243, 1984.

\bibitem{tachibana2018efficiently}
Hideyuki Tachibana, Katsuya Uenoyama, and Shunsuke Aihara,
\newblock ``Efficiently trainable text-to-speech system based on deep
  convolutional networks with guided attention,''
\newblock in {\em International Conference on Acoustics, Speech and Signal
  Processing (ICASSP)}. IEEE, 2018, pp. 4784--4788.

\bibitem{wang2015thchs}
Dong Wang and Xuewei Zhang,
\newblock ``Thchs-30: A free chinese speech corpus,''
\newblock {\em arXiv preprint arXiv:1512.01882}, 2015.

\bibitem{zen2019libritts}
Heiga Zen, Viet Dang, Rob Clark, Yu~Zhang, Ron~J. Weiss, Ye~Jia, Zhifeng Chen,
  and Yonghui Wu,
\newblock ``{LibriTTS: A Corpus Derived from LibriSpeech for Text-to-Speech},''
\newblock in {\em Proc. Interspeech 2019}, 2019, pp. 1526--1530.

\bibitem{maaten2008visualizing}
Laurens Maaten and Geoffrey Hinton,
\newblock ``Visualizing data using {t-SNE},''
\newblock {\em Journal of machine learning research}, vol. 9, no. Nov, pp.
  2579--2605, 2008.

\end{thebibliography}

\end{document}